\documentclass[letterpaper, 10 pt, conference]{ieeeconf}  % Comment this line out if you need a4paper

\IEEEoverridecommandlockouts                              % This command is only needed if 
\usepackage{graphics} % for pdf, bitmapped graphics files
\usepackage{graphicx}
\usepackage{amsmath}
\usepackage[table]{xcolor}

\title{\LARGE \bf
A Tip Mount for Transporting Sensors \\and Tools using Soft Growing Robots
}

\author{
Sang-Goo Jeong$^{1, 2 *}$, Margaret M. Coad$^{1 *}$, Laura H. Blumenschein$^{1}$, Ming Luo$^{1}$, \\Usman Mehmood$^{2}$, Ji Hun Kim$^{2}$, Allison M. Okamura$^{1}$, and Jee-Hwan Ryu$^{3}$
\thanks{This research was partially supported by the project ``Toward the Next Generation of Robotic Humanitarian Assistance and Disaster Relief: Fundamental Enabling Technologies (10069072)," the National Science Foundation (NSF) under Award 1637446, the Air Force Office of Scientific Research under award FA2386-17-1-4658, Toyota Research Institute (TRI), and the NSF Graduate Fellowship Program. TRI provided funds to assist the authors with their research, but this article solely reflects the opinions and conclusions of its authors and not TRI or any other Toyota entity.}% <-this % stops a space
\thanks{*These authors contributed equally to this work.}
\thanks{$^{1}$Department of Mechanical Engineering, Stanford University, Stanford, CA 94305 USA
        {\tt\small \{mmcoad, lblumens, mingluo, aokamura\} @stanford.edu}}%
\thanks{$^{2}$School of Mechanical Engineering, Korea University of Technology and Education, Cheonan-si, Republic of Korea
        {\tt\small \{jsg1215z, umehmood, vgty5678\} @koreatech.ac.kr}}%
\thanks{$^{3}$Department of Civil and Environmental Engineering, KAIST, Daejeon, South Korea. 
        {\tt\small jhryu@kaist.ac.kr}}%        
}

\begin{document}
\graphicspath{{Pictures/}}

\maketitle
\thispagestyle{empty}
\pagestyle{empty}

%%%%%%%%%%%%%%%%%%%%%%%%%%%%%%%%%%%%%%%%%%%%%%%%%%%%%%%%%%%%%%%%%%%%%%%%%%%%%%%%
\begin{abstract}
Pneumatically operated soft growing robots that extend via tip eversion are well-suited for navigation in confined spaces. Adding the ability to interact with the environment using sensors and tools attached to the robot tip would greatly enhance the usefulness of these robots for exploration in the field. However, because the material at the tip of the robot body continually changes as the robot grows and retracts, it is challenging to keep sensors and tools attached to the robot tip during actuation and environment interaction. In this paper, we analyze previous designs for mounting to the tip of soft growing robots, and we present a novel device that successfully remains attached to the robot tip while providing a mounting point for sensors and tools. Our tip mount incorporates and builds on our previous work on a device to retract the robot without undesired buckling of its body. Using our tip mount, we demonstrate two new soft growing robot capabilities: (1) pulling on the environment while retracting, and (2) retrieving and delivering objects. Finally, we discuss the limitations of our design and opportunities for improvement in future soft growing robot tip mounts.
\end{abstract}

%%%%%%%%%%%%%%%%%%%%%%%%%%%%%%%%%%%%%%%%%%%%%%%%%%%%%%%%%%%%%%%%%%%%%%%%%%%%%%%%
\section{Introduction}

Many potential robotic applications require the transport of sensors and tools through confined spaces to explore and interact with the environment. Continuum robots have particular strengths for these types of applications due to their ability both to pass through small apertures and to support their body weight to rise up over obstacles. For example, a camera-equipped snake robot~\cite{Whitman2018} was deployed within a collapsed building for search and rescue after the 2017 Mexico City earthquake. Additionally, a gripper-equipped snake-like mobile robot~\cite{Tanaka2019} has been demonstrated for grasping and retrieving objects and turning a valve in a mock disaster scenario.

\begin{figure}[tb!]
	\begin{center}
		\vspace{0.3cm}
		\includegraphics[width=\columnwidth]{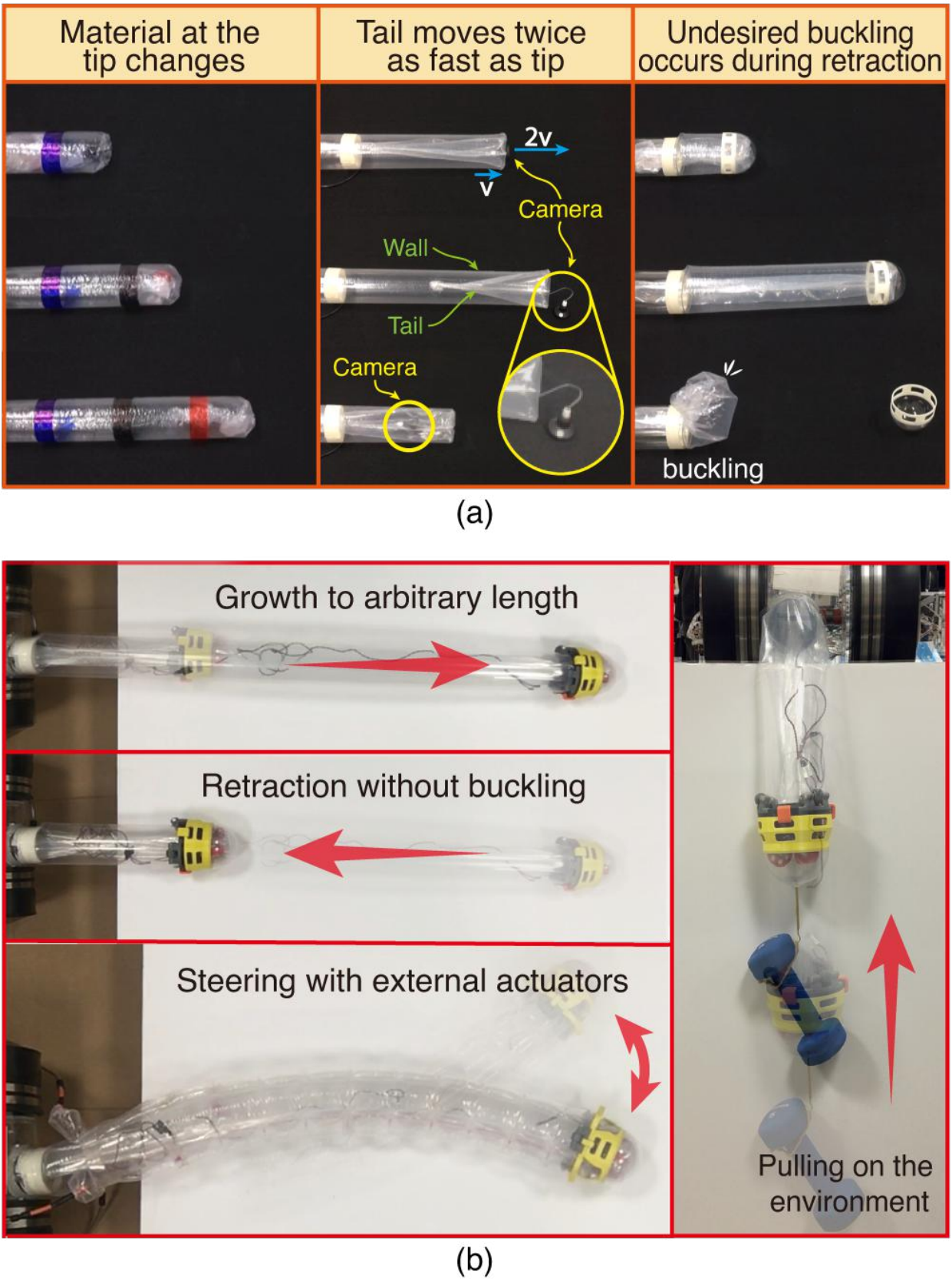}
		\caption{(a) Challenges related to mounting to the tip of soft growing robots. (\textit{left}) The material at the robot tip changes during growth and retraction, so anything affixed to the tip material will not remain at the tip. (\textit{middle}) The inner material (the ``tail") moves at twice the speed of the robot tip relative to the base, so anything packaged within the tail will be ejected during growth and engulfed during retraction. (\textit{right}) When retracted from the base, soft growing robots often undergo undesired buckling, leading to lack of control of their motion and force. (b) Our current tip mount design successfully surmounts these challenges. Our design remains at the robot tip during growth to arbitrary lengths as well as retraction and incorporates a device~\cite{CoadRetraction2020} to retract the robot without undesired buckling. Like previous tip mounts, our design does not interfere with steering of the robot body using external actuators. Our tip mount can also apply significant pulling forces to the environment.
		}
		\vspace{-0.7cm}
		\label{fig:Main Figure}
	\end{center}
\end{figure}

Another type of continuum robot that is well-suited for navigation in confined spaces is a pneumatically driven soft growing robot~\cite{mishima2006development,HawkesScienceRobotics2017}. Unlike typical snake-like robots, this type of robot ``grows" from a fixed base by transporting new material through its body and everting the material at its tip, driven by internal air pressure. These robots are particularly good at traversing long, tortuous paths over varied terrain due to their ability to grow to arbitrary lengths from a fixed base~\cite{HawkesScienceRobotics2017,CoadRAM2020}; propel their tip forward without relative movement between their body and the environment~\cite{HawkesScienceRobotics2017,blumenschein2017modeling}; conform to the shape of their surroundings via their natural softness~\cite{Greer2018ICRA,Haggerty2019IROS}; shrink in diameter to pass through apertures smaller than their body diameter~\cite{HawkesScienceRobotics2017,CoadRAM2020,Nakamura2018AIM}; support their own body weight to move over obstacles and across gaps~\cite{HawkesScienceRobotics2017,CoadRAM2020}; and controllably direct their tip in combination with growth~\cite{HawkesScienceRobotics2017,CoadRAM2020,Blumenschein2018Robosoft,greer2019soft}.

% The direction of movement of the robot tip can be actively controlled in three degrees of freedom through the addition of various steering mechanisms that lengthen or shorten one side of the robot body relative to the opposite side, enhancing their ability to navigate their environment. 

Adding the ability to transport sensors and tools at the tip of soft growing robots during movement and environment interaction would greatly expand their usefulness in the field. Mounting cameras and other sensors at the tip of the robot would enable information gathering tasks (e.g., collecting data in an unknown environment), while mounting tools such as grippers at the tip of the robot would allow delivery and retrieval of objects in the environment (e.g., transporting supplies to a trapped disaster victim or retrieving items from a confined space) and application of force to the environment (e.g., to turn a valve or open a door).

However, mounting to the robot tip in a reliable manner is a key challenge for soft growing robot design. Unlike many continuum robots, the material at the tip of the soft growing robot continually changes as the robot grows and retracts. Thus, anything affixed to the robot body material that is currently at the robot tip will not remain at the tip as the robot grows or retracts (Fig.~\ref{fig:Main Figure}(a, \textit{left})). As the robot grows, the current tip material becomes part of the stationary outer robot body (the ``wall"), and as the robot retracts, the tip material becomes part of the inner robot body (the ``tail"), which moves towards the base at twice the speed of the robot tip (Fig.~\ref{fig:Main Figure}(a, \textit{middle})). Therefore, sensors and tools must move relative to the robot body material to remain at the robot tip and cannot be permanently attached to the body wall through simple means like tape or glue. %and other mechanisms of applying the force required to keep it at the robot tip must be explored. 
This challenge is shared by other tip-growing robots~\cite{tsukagoshi2011tip,sadeghi2017toward} and everting toroidal robots~\cite{orekhov2010mechanics,orekhov2010actuation}.

% A successful soft growing robot tip mount must stay on the robot tip for all desired movement and environmental interaction of the robot. Such a mount could be used to carry sensors or tools to physically interact with the environment (Fig.~\ref{fig:Main Figure}).

Various designs for tip mounts for soft growing robots have previously been developed~\cite{mishima2006development, HawkesScienceRobotics2017,CoadRAM2020,greer2019soft,luong2019eversion,Stroppa2020Human} and will be described in detail in Sections~\ref{background} and~\ref{discussion}. One design~\cite{CoadRAM2020} carried a camera during deployment of a soft growing robot for exploration of an archeological site, and another design~\cite{Stroppa2020Human} carried a gripper and transported lightweight objects during an object delivery task. However, none of the previous designs is able to remain at the robot tip during retraction as well as growth, transmit significant pulling forces from the robot body to the environment, and function consistently at an arbitrary robot body length. An additional shortcoming of all previous tip mount designs is that none of them incorporates a device to retract the soft growing robot without undesired buckling (and thus lack of control) of its body (Fig.~\ref{fig:Main Figure}(a, \textit{right})), a problem solved in our previous work~\cite{CoadRetraction2020} and described in detail in Section~\ref{retraction}. %Tip mounts that achieve all of these features will be crucial for tasks in the field that require controlled retraction, pulling on the environment, and growth to arbitrary lengths, such as object retrieval and environment force application tasks that occur far from the robot base.

Improving upon previous work, we present a new tip mount for soft growing robots (Fig.~\ref{fig:Main Figure}(b)) that (1) reliably remains at the robot tip during retraction and growth, (2) transmits pulling forces from the robot to the environment, (3) functions at an arbitrary robot body length, and (4) incorporates a device to retract the soft growing robot without undesired buckling. Additionally, like other tip mount designs, our tip mount allows steering of the robot tip using external actuators such as series pouch motors~\cite{CoadRAM2020}.

\begin{figure*}[htb]
    \centering
    	\vspace{0.3cm}
    \includegraphics[width=\textwidth]{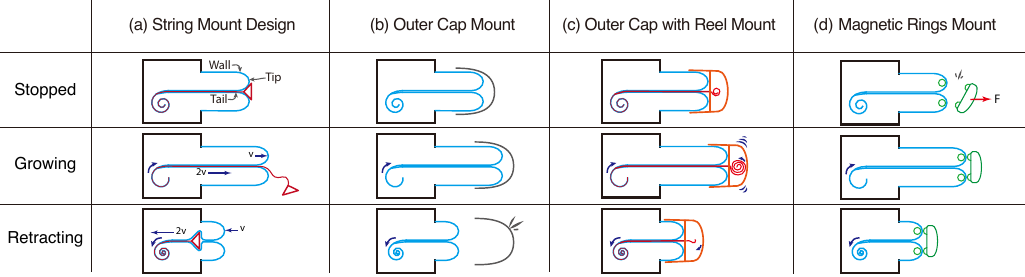}
    \caption{Function of previous soft growing robot tip mount designs during growth and retraction. Each design has benefits and at least one limitation. (a) The string mount~\cite{HawkesScienceRobotics2017,greer2019soft} implemented on a spooled tail robot gets ejected during growth and engulfed during retraction. (b) The outer cap mount~\cite{CoadRAM2020} remains at the robot tip during growth but falls off during retraction. (c) The outer cap with reel mount~\cite{mishima2006development} remains at the robot tip during both growth and retraction, but the reel at the tip increases in size as the robot body grows, and it may become too large or heavy when the robot reaches a long length. (d) The magnetic rings mount~\cite{luong2019eversion,Stroppa2020Human} remains at the robot tip during growth and retraction but is susceptible to falling off due to forces from the environment.}
    \vspace{-0.3cm}
    \label{fig:previousresearchfig}
\end{figure*}

\section{Previous Design Summary} \label{background}
Motivated by useful tasks that require transport of sensors and tools at the tip of soft growing robots, four different tip mount designs have previously been developed. Here, we present a summary of each previous design and discuss each design's benefits and limitations related to the goals of (1) remaining at the robot tip during retraction and growth, (2) transmitting pulling forces to the environment, and (3) functioning at an arbitrary robot body length (Fig.~\ref{fig:previousresearchfig}). We also explain the problem of undesired buckling during retraction (a limitation of soft growing robots that affects all previous designs) and summarize our previous work on a device to retract soft growing robots without undesired buckling, a building block towards our current design. Other attributes of previous designs, including the weight they add to the robot tip and whether they encumber the natural abilities of soft growing robots to move through constrained environments, are discussed in Section~\ref{discussion}.

\subsection{String Mount}
One previous tip mount design was used in \cite{HawkesScienceRobotics2017,greer2019soft} to transport a camera at the tip of a soft growing robot during laboratory demonstrations of the robot's navigation ability. This relatively simple design ties a string to the tip mount (which can be the sensor or tool itself) and uses the robot tail as a conduit for the string to pass from the robot tip to the base. Thus, the base can be the grounding point for the force to keep the mount at the tip. However, as the robot body is pressurized, the tail squeezes around the string, forcing the string to move with the tail. Since the material of the tail moves twice as fast as the tip relative to the base, the string and tip mount are ejected from the robot during growth and engulfed into the body during retraction (Figs.~\ref{fig:Main Figure}(a, \textit{middle}) and \ref{fig:previousresearchfig}(a)). In \cite{HawkesScienceRobotics2017,greer2019soft}, to overcome this issue during growth, airflow was added inside the tail, and the string was pulled back from the base to keep the mount at the tip, which required the tail material to be stored straight rather than on a spool. This limits the potential length change and does not allow growth to arbitrary lengths. Also, this design can only remain at the robot tip during growth, not retraction, since the string can only apply tensile, not compressive forces. One benefit of this design is the physical connection back to the base through the string, which can be used to transmit significant pulling forces to the environment.

%A second limitation is that this design cannot remain at the robot tip during growth when the tail is stored on a spool in the base (a storage method that enables growth to arbitrary lengths). Instead, the robot must have a straight tail, in which case it is unable to shorten to less than half of its fully grown length, which is a limitation for applications where there is little space at the robot base.

\subsection{Outer Cap Mount}
A second design was used in \cite{CoadRAM2020} to transport a camera on a robot deployed in the field for exploration of an archeological site. This design is also quite simple and uses a rigid cap that fits over the outside of the robot tip and is pushed along by the robot's growing force. Unlike the string mount design, this design does not have a direct connection to the robot base and relies on friction between the inside of the cap and the outside of the robot body wall to keep the mount at the tip (Fig.~\ref{fig:previousresearchfig}(b)). The size of the outer cap can be varied relative to the robot body size, where smaller caps provide higher normal and thus frictional forces. 

Unlike the string mount, this design functions for arbitrary robot body lengths and can be used when the robot is stored in a reel at the base, enabling enormous length change from a small initial form factor. However, this design has limited ability to remain at the robot tip during retraction. While the friction between the outer cap and the robot body wall holds the cap on the tip during growth, it does not have the same effect during retraction. Instead, the cap remains behind as the robot retracts within it. This design also cannot transmit significant pulling forces to the environment. The frictional forces between the outer cap and the robot body wall determine the largest pulling forces that can be transmitted to the environment, and a deterrent to choosing a design with high frictional forces is the resulting increase in the pressure required to begin growth of the robot, as discussed for our current design in Section~\ref{pressure_grow}.%, since the added friction makes relative movement between the cap and the wall more difficult. This means that a higher internal pressure is required to begin growth and to grow at a given speed. The burst pressure of the robot body provides an upper limit on the allowable internal pressure, and thus there is an upper limit on the frictional attachment force of this mount such that growth is still possible at the desired speed.

\subsection{Outer Cap with Reel Mount}
A third tip mount design was used in \cite{mishima2006development} to transport a camera during a laboratory demonstration of robot growth. This design combines features of the string mount and outer cap mount, with an outer cap (on which the sensor/tool can be mounted) containing a motorized reel attached to a string running internal to the tail. During growth, as the string ejects from the tip, the motor actively reels in the slack, keeping the cap at the tip, while during retraction, the motor lets out slack to feed the string into the tail (Fig.~\ref{fig:previousresearchfig}(c)). This successfully keeps the mount at the tip during retraction, but does not solve the problem of buckling during retraction. Like with the string mount, the force to hold the mount at the tip comes from the connection of the string back to the base, so the outer cap stays at the robot tip during retraction and growth, and this connection back to the base can be used to transmit significant pulling forces to the environment.

%One benefit of this design is that, unlike either the string mount design or the outer cap design, it successfully remains at the robot tip during retraction. 

%A second benefit is that, like the string mount design, this design's physical .

This design is limited in that, unlike the outer cap design, it does not function at arbitrary robot body lengths. As the length of the robot increases, the reel inside the mount needs to grow to hold the extra string, limiting the length of the robot based on the size of the tip mount.

\subsection{Magnetic Rings Mount}
The fourth previous tip mount design was used in \cite{luong2019eversion} to transport a camera during a laboratory demonstration of growth of a water-filled robot, as well as in \cite{Stroppa2020Human} to transport a gripper for completion of a pick-and-place task in a laboratory environment. This is the first of the tip mount designs to place part of the mount inside the pressurized area of the robot body. This design consists of a ring inside the tip of the robot and another ring (on which the sensor/tool is mounted) outside the tip of the robot. The two rings are held together, and at the tip, using magnetic rollers that roll along the robot body material during growth and retraction, allowing the material to pass between the two halves of the mount in a low-friction manner (Fig.~\ref{fig:previousresearchfig}(d)).

This design is able to remain at the robot tip during retraction as well as growth, independent of body length, provided that the outer ring is large enough in diameter that it will not get engulfed into the robot body. However, like the outer cap mount, the pulling forces that can be transmitted to the environment are limited. In this case, the magnetic force between the two halves of the mount, limited by the strength of the magnets used, is the upper bound on pulling forces.

%A second benefit of this design is that, like the outer cap mount, it functions consistently at arbitrary robot body lengths, since it does not rely on the use of a string internal to the robot tail.

\subsection{Retraction Without Buckling} \label{retraction}
A limitation with soft growing robots is the tendency of their body to undergo undesired buckling when retraction is attempted by pulling on the tail from the robot base (Fig.~\ref{fig:Main Figure}(a, \textit{right})), causing a lack of control of body motion and force applied to the environment. This limitation is likely the reason why none of the previous tip mount demonstrations showed any significant retraction of the soft growing robot body. Our previous work \cite{CoadRetraction2020} analyzes the problem of undesired buckling during retraction and its dependence on robot length, curvature, and pressure. In summary, buckling occurs because the force required to invert the soft robot body is independent of length, but the force required to buckle the soft robot body decreases with length. Thus, regardless of internal pressure and curvature, soft growing robots will always buckle rather than inverting above a certain length when the tail is pulled from the base.

Our previous work presents a solution to buckling in the form of a ``retraction device," which sits at the robot tip and uses motor-driven rollers to apply force on the tail, grounded to the robot tip. Applying force from the tip rather than from the base makes the effective robot body length zero, enabling retraction without undesired buckling at any robot length, pressure, and curvature.

Both retraction devices and tip mounts must sit at the robot tip, so it makes sense to combine them in a tip mount for tasks that require retraction of the soft growing robot.

% \subsection{Previous Design Summary} \label{prevSummary}
% A summary of the past designs and their capabilities can be found in Table~\ref{tab:previousdesign}. These previous tip mount designs identified a number of distinct strategies for staying attached to a moving point. The designs demonstrate a large range of potential mounting locations, including within the tail, external to the robot, and within the robot body. As well, many or all the designs share some features in common, including a surface at the tip to be pushed forward as the robot grows. While none of these designs completely meets the goal of staying mounted during any potential robot actuation or external force, the designs overall give a guide to a fully successful tip mount. However, none of the designs incorporates a way to mitigate buckling of the soft robot body during retraction. Another important feature missing from the existing designs, with the potential exception of the string mount, is the ability to resist tension loads, which will open up the applications of soft growing robots with tip mounts to include more manipulation tasks.

\section{Proposed Tip Mount Design} \label{design}
Our soft growing robot tip mount design aims to overcome the limitations of previous designs in completing tasks that require object retrieval and environment force application. We present a tip mount that (1) remains at the robot tip during growth and retraction, (2) transmits pulling forces between the robot body and the environment, (3) functions at an arbitrary robot body length, and (4) incorporates a retraction device to prevent buckling during retraction.

Our design combines concepts from the outer cap design~\cite{CoadRAM2020} and the magnetic rollers design~\cite{luong2019eversion,Stroppa2020Human}, and it incorporates a retraction device~\cite{CoadRetraction2020}. The retraction device sits inside the pressurized area at the robot tip, similar to the inner ring of the magnetic rollers design. An outer cap outside the robot tip provides a mounting location for sensors and tools. A key improvement of our current design upon the magnetic rollers design is the attachment mechanism between the outer and inner parts of the mount. Rather than relying on magnetic force to hold the two parts together, our design employs a rolling interlock where the outer cap hooks around the inner retraction device so that the two pieces cannot physically be separated without breaking. Each piece has passive rollers at the connection point so that the material of the soft robot body wall can pass between them in a low friction manner. While this tip mount does not have a physical connection back to the base like the string mount~\cite{HawkesScienceRobotics2017,greer2019soft} and outer cap with reel mount~\cite{tsukagoshi2011tip}, it can transmit significant forces through the rolling interlock, which is grounded to the robot tip using the retraction device. These forces hold the mount at the robot tip and allow significant pulling forces to be transmitted to the environment: from the base, along the robot body, and then through the mount. The following subsections describe in detail the mechanical design and control of our current tip mount design.

\begin{figure}[t]
	\centering
	\vspace{0.3cm}
		\includegraphics[width=0.85\columnwidth]{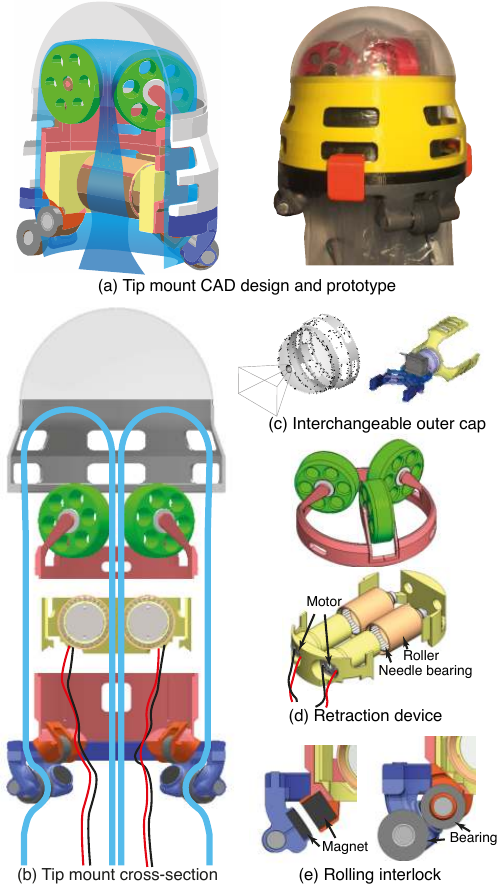}
		\caption{Our current tip mount design. (a) (\textit{left}) CAD rendering and (\textit{right}) photo of the tip mount. (b) The tip mount consists of three parts: (c) outer cap to mount (\textit{left}) sensors or (\textit{right}) tools, (d) retraction device including (\textit{top}) passive rollers to decrease friction with the robot tip and (\textit{bottom}) motor-driven rollers to apply retraction forces on the robot tail, and (e) a rolling interlock using (\textit{left}) magnets and (\textit{right}) bearings to hook the inner and outer parts together.
		}
		\vspace{-0.3cm}
		\label{fig:exploded}
\end{figure}

\subsection{Mechanical Design}
Our current tip mount design (Fig.~\ref{fig:exploded}) consists of three main components: (1) an outer cap for mounting, (2) a retraction device to allow retraction without buckling, and (3) a rolling interlock to attach them together.

\paragraph{Outer Cap}
The outer cap (Fig.~\ref{fig:exploded}(c)) provides a mounting location for sensors and tools and a place for the robot body to push on and propel the tip mount forward. The outer cap design should not impede the function of the steering actuators used to direct the tip of the robot. In our implementation, we used three series pouch motor steering actuators, which shorten when pressurized and are attached circumferentially around the robot body, as in~\cite{CoadRAM2020}, so we added three cutouts in the cap to give the actuators room to inflate and deflate without affecting the fit on the robot body.

\paragraph{Retraction Device}
The retraction device (Fig.~\ref{fig:exploded}(d)) is as presented in~\cite{CoadRetraction2020}, with stronger motors and a different tip grounding mechanism. During retraction, two motors (3485, Pololu Corporation, Las Vegas, NV) drive rollers coated in high friction material (Non-Slip Reel, Dycem Corporation, Bristol, UK) that squeeze the tail material and pull it toward the base. Three passive rollers at the top of the device apply a reaction force to the robot tip while allowing the material to move easily around the tip. During growth, the active rollers are driven to move tail material towards the robot tip, and the material pushes on the outer cap instead of contacting the passive rollers. Aside from the high friction material coating the actively driven rollers, all other parts are designed to reduce unnecessary friction with the soft robot body material.

\paragraph{Rolling Interlock}
The rolling interlock (Fig.~\ref{fig:exploded}(e)) consists of three matching sets of roller-magnet units, placed circumferentially around the base of the outer cap and the base of the retraction device, such that the series pouch motors lie in between. Each roller-magnet unit has a passive roller with a disk-shaped magnet on either side. Only the rollers contact the wall material, which must slide between the pairs of disk-shaped magnets (separated by a small space) as the robot grows or retracts. The rollers can transmit high forces across the membrane, and the magnets prevent relative tilting or rotation (and thus separation) between the outer cap and retraction device. We use separate rollers and magnets, as opposed to magnetic rollers, since disk magnets can provide higher magnetic force.

%The magnets are attached diagonally to restrain movement in any direction, and there is a small space between each pair of magnets through which the wall material can travel. The proper spacing of the magnets is highly dependent on the robot body diameter used, and even on the pressure inside the robot body, as the diameter can change slightly with pressure. Roller-magnet units were selected rather than magnetic rollers due to the ability to apply a higher magnetic force using a disk magnet configuration. Three roller-magnet units were chosen to contact the soft robot body wall between the three series pouch motor actuators. To reduce friction between the retraction device and the inside of the robot body wall, the outer edges of the passive rollers on top of the retraction device and the outer edges of the passive rollers in the rolling interlock are aligned with each other.

\subsection{Control}
We used the base, joystick, and steering control algorithm presented in \cite{CoadRAM2020} to steer the soft robot body by coordinating pressures (0-14~kPa) in the three series pouch motor actuators placed circumferentially around the body of the robot. Additionally, we developed a method of coordinating the voltages sent to the base motor and the retraction device motors to allow growth and retraction without building up slack in the tail or buckling the body. For simplicity, we used open-loop voltage control of the motors with no position sensing.

\paragraph{Growth}
During growth, the pressure in the soft robot body was set (using the joystick) higher than needed to grow at the desired speed (0-17~kPa), the motor in the base was backdriven to let out the tail material without building up slack, and the motors in the retraction device (which are not backdrivable) were controlled to release the material at the desired speed. To achieve this, we set the voltage of the motor in the robot base to offset static friction in the motor (3.5~V), and the voltage of the motors in the retraction device based on the joystick input to be between 2.4~V and 15~V. Using this control method, the soft robot body with our tip mount attached can grow at a maximum speed of 5~cm/s.

\paragraph{Retraction}
During retraction, the pressure in the soft robot body was set (using the joystick) as low as possible while keeping the robot body pressurized (approximately 7~kPa), to allow easy sliding of the robot body material between the magnets while limiting the retraction force. The motors in the base were run with enough force to take in the slack in the tail but not to buckle the body, and the motors in the retraction device provided the rest of the necessary force to retract. To achieve this, we set the voltage of the motor in the robot base to the highest voltage before the straight robot body began to buckle at the retraction pressure (9.4~V), and the voltage of the motors in the retraction device based on the joystick, between 2.4~V and 15~V. Using this control method, the soft robot body with our tip mount attached is able to retract at the same maximum speed as growth: 5~cm/s.

\section{Characterization} \label{character}
We conducted two experiments to characterize the capabilities of a soft growing robot with our tip mount design. The first quantifies the effect of the additional friction due to the tip mount on the pressure required to grow the robot. The second quantifies the pulling force that can be transmitted to the environment. For both experiments, we explore how the design of different portions of the device affects the robot's capabilities. Throughout this paper, the robot body was made using a tube of low-density polyethylene (LDPE) plastic, a material easy to use for quick prototyping, with inflated diameter 8.5~cm and wall thickness 60~$\mu$m. 

\subsection{Minimum Pressure to Grow}\label{pressure_grow}
The minimum pressure required to begin growth of a soft growing robot is an important predictor of its capabilities. The internal pressure can be set higher than the minimum pressure required to begin growth, and, if the tail is free to move forward, the additional pressure will either make the robot grow faster~\cite{HawkesScienceRobotics2017,blumenschein2017modeling} or apply more force at its tip~\cite{mishima2006development,godaba2019payload}, up to the buckling load of the robot body. The maximum internal pressure is limited by the burst pressure of the soft robot body. The addition of our tip mount adds friction between the body material and the mount and between the mount and the environment, increasing the minimum pressure required to grow. This decreases the maximum growth speed and the maximum pushing force that can be applied before bursting the soft robot body.

To understand the effect of our tip mount on the pressure to grow (and thus the growth speed and pushing force capability), we conducted growth tests horizontally on a foam board floor with different parts of our device installed. We slowly increased the pressure with slack in the tail and observed the minimum pressure at which growth occurred.

\begin{figure}[t]
	\centering
	\vspace{0.3cm}
	\includegraphics[width=\columnwidth]{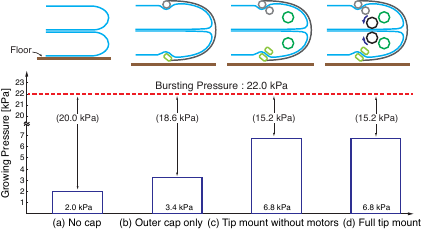}
	\caption{The minimum pressure required to begin growth as compared to the soft growing robot burst pressure when different portions of our current tip mount are attached to the robot body: (a) no tip mount, (b) only the outer cap, (c) the retraction device and the outer cap are attached, but the motors and active rollers are removed, and (d) the full tip mount is attached. The addition of the tip mount decreases the difference between the minimum pressure to grow and the burst pressure by only 24\%, meaning that the impact on relevant robot capabilities is relatively small.
	}
	\vspace{-0.3cm}
	\label{fig:minpressure}
\end{figure}

The results are shown in Fig.~\ref{fig:minpressure}. The soft robot body without any tip mount (Fig.~\ref{fig:minpressure}(a)) requires 2~kPa to begin growing, due to the forces needed to turn the soft robot body inside out at its tip. The addition of the outer cap (Fig.~\ref{fig:minpressure}(b)) increases the required pressure to 3.4~kPa. This represents the friction as the wall of the soft robot body slides against the outer cap and the outer cap slides on the floor. Adding the inner part without the motors and rollers (Fig.~\ref{fig:minpressure}(c)) increases the pressure to 6.8~kPa, due to the friction at the point of contact between the retraction device and the outer cap, where there is some sliding between the magnets and wall material. Finally, with the addition of the motors and rollers (Fig.~\ref{fig:minpressure}(d)), the robot still requires 6.8~kPa to grow, since the rollers do not slide on the tail.

The largest friction increase occurs at the bearing and magnet interface, so improving the design there would have the largest impact. Other locations to improve friction are between the tip mount and the environment and between the outer cap and the robot wall. The experimentally determined burst pressure of the soft robot body is 22.0~kPa, so the additional friction of this device decreases the available range of pressure above the minimum growth pressure by only 24\% (from 20.0~kPa to 15.2~kPa). %These results indicate that the most important location to decrease friction in the device is in the bearings and magnets interface between the outer cap and the retraction part. Additional locations that would have a smaller impact on decreasing the minimum growth pressure are between the outer cap and the environment, as well as between the outer cap and the wall. The experimentally determined burst pressure of the soft robot body is 22.0~kPa, so the additional friction of this device decreases the available range of pressure above the minimum growth pressure by only 24\% (from 20.0~kPa to 15.2~kPa).

\subsection{Maximum Pulling Force}
An important goal of our tip mount design is to allow transmission of significant pulling forces to the environment. Soft growing robots are much weaker in pushing and side loading than traditional engineering materials, so their ability to pull on the environment is vital. The force these robots can support in compressive or transverse loading before buckling can be calculated using inflated beam models~\cite{fichter1966theory,levan2005bending,leonard1960structural,comer1963deflections} and decreases with length. However, the force these robots can pull is length-independent and depends only on the mechanical properties of their body material. Because soft growing robots can grow to arbitrary lengths, harnessing this pulling capability is key to useful environment interaction and applications such as turning a valve, opening a door, or retrieving items from a confined space.

\begin{figure}
    \centering
    \vspace{0.3cm}
    \includegraphics[width=\columnwidth]{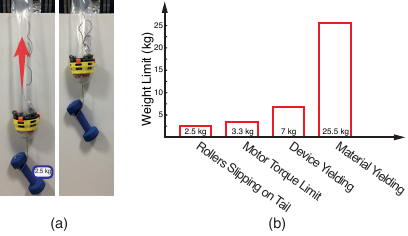} 
    \caption{Vertical weight pulling limit for a soft growing robot with our tip mount attached, based on various factors. (a) Using our current tip mount, the soft growing robot can pull a weight of up to 2.5~kg while retracting. (b) Four factors contribute to the maximum pulling force. The rollers slipping on the tail is currently the limiting factor, but all factors could potentially be improved in future designs.}    \label{fig:lifting}
    \vspace{-0.3cm}
\end{figure}

To quantify the pulling ability of a soft growing robot with our tip mount, we hung the robot vertically downward and attached a weight to a hook on the tip mount. We then attempted to retract the weight with the robot and increased the weight until failure (2.5~kg) (Fig.~\ref{fig:lifting}(a)). The robot's pulling force could be limited by the frictional force the retraction device rollers transmit to the tail, the torque of the retraction device motors, the breaking strength of the tip mount material, or the yielding strength of the robot body material. We calculated or measured the pulling limits based on each of these factors, as described in the following subsections and shown in Fig.~\ref{fig:lifting}(b).

\paragraph{Rollers Slipping on Tail}
The force applied by the retraction device on the tail of the robot is transmitted at the connection point between the motor-driven rollers and the tail. The maximum measured force that can be applied by the rollers on the tail before slip occurs is 5~kg. Frictional losses in the tip mount, the need to counter the internal pressure of the soft robot body, and the weight of the tip mount itself (0.5~kg), further limit the weight that our tip mount can lift to 2.5~kg. This is the limiting factor in force that the soft growing robot can currently pull, but increasing the force the rollers can apply before slipping would increase this limit.

\paragraph{Motor Torque Limit}
The motors in the device have a gearbox torque limit of 5~kg$\cdot$cm, so the maximum torque that the two motors can withstand together is 10~kg$\cdot$cm. With a roller radius of 3~cm, this results in a motor torque pulling limit of 3.3~kg. If the roller slip force is increased, the use of stronger motors could increase the force limit. %However, due to frictional losses in the device and the need for the motors to counteract the internal pressure of the soft robot body and to lift the weight of the device, the actual maximum weight that could be lifted with these motors is lower than this value, but it could not be experimentally determined, because the rollers began slipping on the tail before the motors reached their limit. 

\paragraph{Device Yielding}
We applied an increasing force until the tip mount came apart. When a load of 7~kg was applied, the outer cap flexed and broke, which could be improved by reinforcing the design and switching to a stronger material than the 3-D printed PLA used.

\paragraph{Material Yielding}
To calculate the pulling force limit due to yielding of the soft robot body material, we first experimentally determined the yield stress of the material by increasing the pressure until the robot body burst ($P = 22.0$~kPa). Using the equation for hoop stress in a thin-walled cylinder (the highest-stress direction), we can calculate the yield stress of the material:
\begin{equation}
    \sigma=\frac{Pr}{t},
\end{equation}
where $P$ is the pressure inside the robot, $r$ is the tube radius ($r = 4.25$~cm), $t$ is the material thickness ($t=60~\mu$m) and $\sigma$ is the hoop stress at yield, which was calculated to be 15.6~MPa. To reach material yielding with a pulling force instead, we need a force of 25.5~kg. This force limit could be improved by using a material with a higher yield stress, or by increasing the cross-sectional area of the body wall, though this would also affect burst pressure and growth pressure.

Overall, with our tip mount implementation, we are for the first time able to apply significant pulling force to the environment while retracting without buckling. The 2.5~kg of pulling force that we are able to transmit to the environment is only 10\% of the 25.5~kg of potential force transmission of the robot body material. This is encouraging, as it indicates there is room to better implement our design (e.g., through rollers capable of transmitting more force before slipping, and higher-torque motors) to exert even higher pulling forces on the environment.

\section{Demonstration} \label{demo}
With our current tip mount, soft growing robots can for the first time grow to arbitrary lengths and retract without buckling while transporting sensors and tools at the robot tip, all while steering with external actuators such as series pouch motors. In addition to pulling forces, our tip mount can support compression loads, allowing the robot to push objects up to the buckling load and burst pressure of the soft robot body. These capabilities will greatly enhance the usefulness of soft growing robots for tasks in the field.

\begin{figure*}[ht]
    \centering
    \vspace{0.3cm}
	\includegraphics[width=\textwidth]{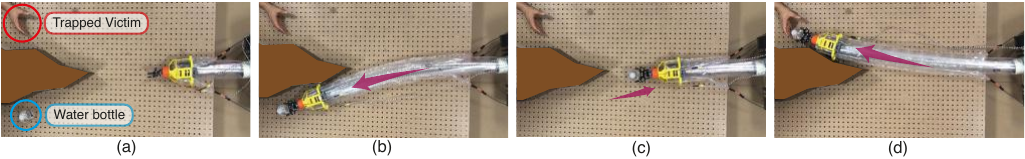}
	\caption{Demonstration of a new capability of object retrieval and delivery made possible by our current tip mount design. (a) In a mock disaster scenario, the soft growing robot (b) grows and steers to pick up a water bottle, (c) retracts with the water bottle, and (d) grows and steers and places the object in the trapped victim's hand. 
	}
	\label{fig:demo}
	\vspace{-0.3cm}
\end{figure*}

To demonstrate the usefulness of these capabilities, we show a simple object retrieval and delivery task that involves growth, retraction, and steering using a gripper-equipped robot. In Fig.~\ref{fig:demo}(a and b), the robot grows and steers along the floor, approaching and grabbing the water bottle. It then retracts without buckling to move around an obstacle (Fig.~\ref{fig:demo}(c)) before growing and steering in the opposite direction, to head to the hand of the trapped person (Fig.~\ref{fig:demo}(d)).

While this demonstration is not necessarily representative of a real disaster scenario, it showcases the robot capabilities to pull/push objects while growing, retracting, and steering through an environment, which are made possible for the first time with our tip mount design.

\section{Discussion} \label{discussion}
Unlike the four previous soft growing robot tip mount designs, our current tip mount design (1) remains at the robot tip during growth and retraction, (2) transmits pulling forces to the environment, (3) functions at arbitrary robot body lengths, and (4) incorporates retraction without buckling. However, our design has some limitations that are not shared by all previous tip mount designs: it (1) adds significant weight to the robot tip, (2) slides relative to the environment, and (3) does not allow body shrinking through apertures. Table~\ref{tab:previousdesign} summarizes the capabilities and limitations of all five tip mount designs discussed in this paper.

The additional weight of our current tip mount (0.5~kg) comes primarily from the two motors of the retraction device, though the bearings and magnets that make up the rolling interlock also contribute. For the soft robot body in this paper, this additional weight greatly decreases the maneuverability of the robot tip, making it impossible for the series pouch motor actuators to lift the tip against gravity, and undercutting the natural ability of soft growing robots to support their own body weight over obstacles and across gaps. The limitation of extra weight is shared by the outer cap with reel design, which also contains a motor in the tip mount, while the string mount, outer cap mount, and magnetic rings mount keep the additional weight to a minimum.

The outer cap in our design covers part of the robot body wall and, therefore, slides relative to the environment during growth and retraction. Thus, unlike with soft growing robots without a tip mount, where only internal friction limits growth \cite{blumenschein2017modeling}, here, friction between the environment and outer cap factors into the force required to grow or retract the robot. This limitation is shared by the outer cap mount and the outer cap with reel mount. The magnetic rings and string mount avoid this limitation by only connecting to the tail or the robot tip, and not the wall.

Lastly, the fact that our current tip mount has a rigid outer cap means that it cannot allow the robot body to deform and pass through apertures smaller than its body diameter, another of the natural strengths of soft growing robots. This limitation is also shared by the outer cap mount and the outer cap with reel mount. The magnetic rings mount allows some body shrinking, and the string mount is only limited by the size of the sensor or tool.%, but if the mount is too small, it will tend to be engulfed during retraction. %Here again, the string mount wins out, because it has no extra size besides that of the sensor or tool itself, which is held in the center of the robot tip, allowing body shrinking down to the size of the sensor or tool.

Development of the ideal soft growing robot tip mount design that meets all of the desired capabilities in Table~\ref{tab:previousdesign} is still an open research question, but different tip mounts could work for different applications. For some applications, the limitations of the retraction device and rolling interlock may be worthwhile for the added benefits of controlled retraction and significant pulling force transmission.%, and the loss of some natural soft growing robot capabilities may not matter as much, or vice versa.

An additional consideration for soft growing robot tip mounts not discussed thus far is power and signal transmission to and/or from the sensor or tool at the robot tip. For some applications, wireless signal transmission and battery power may work, but for other applications, a wired connection is crucial. Various methods of passing a wire from the robot base to the tip mount have been developed, including passing a wire inside the tail~\cite{mishima2006development,HawkesScienceRobotics2017,greer2019soft}, inside a self-sealing pocket outside the robot body~\cite{CoadRAM2020}, or entirely outside the robot body~\cite{Stroppa2020Human}. The development of methods to transmit power and signal to the robot tip is an open research question.

\begin{table}[tb]
    \vspace{0.3cm}
    \caption{Capabilities and Limitations of Soft Growing Robot Tip Mount Designs}
    \centering
     \fontsize{6.8}{8}\selectfont
    \begin{tabular}{|p{2.2cm}|p{0.7cm}|p{0.7cm}|p{0.7cm}|p{0.7cm}|p{0.7cm}|}
    \hline
       & String \cite{HawkesScienceRobotics2017,greer2019soft} & Outer Cap \cite{CoadRAM2020} & Outer Cap with Reel \cite{mishima2006development} & Magnetic Rings \cite{luong2019eversion,Stroppa2020Human} & Current Design\\ \hline\hline
    Remains at the robot tip during growth? &\cellcolor{yellow!30!}Some- times &\cellcolor{green!30!}Yes &\cellcolor{green!30!}Yes &\cellcolor{green!30!}Yes &\cellcolor{green!30!}Yes \\    \hline
    Remains at the robot tip during retraction? &\cellcolor{red!30!}No &\cellcolor{red!30!}No &\cellcolor{green!30!}Yes &\cellcolor{green!30!}Yes &\cellcolor{green!30!}Yes \\    \hline
    Can transmit pulling forces? &\cellcolor{green!30!}Yes &\cellcolor{yellow!30!}Some- what &\cellcolor{green!30!}Yes &\cellcolor{yellow!30!}Some- what &\cellcolor{green!30!}Yes \\    \hline
    Functions at arbitrary robot body lengths? &\cellcolor{red!30!}No &\cellcolor{green!30!}Yes &\cellcolor{red!30!}No &\cellcolor{green!30!}Yes &\cellcolor{green!30!}Yes \\    \hline
    Incorporates retraction without buckling? &\cellcolor{red!30!}No &\cellcolor{red!30!}No &\cellcolor{red!30!}No &\cellcolor{red!30!}No &\cellcolor{green!30!}Yes \\    \hline\hline
    Adds minimal weight to the robot tip? &\cellcolor{green!30!}Yes &\cellcolor{green!30!}Yes &\cellcolor{red!30!}No &\cellcolor{green!30!}Yes &\cellcolor{red!30!}No \\    \hline
    Avoids sliding relative to the environment? &\cellcolor{green!30!}Yes &\cellcolor{red!30!}No &\cellcolor{red!30!}No &\cellcolor{green!30!}Yes &\cellcolor{red!30!}No \\    \hline
    Allows body shrinking through apertures? &\cellcolor{green!30!}Yes &\cellcolor{red!30!}No &\cellcolor{red!30!}No &\cellcolor{yellow!30!}Some- what &\cellcolor{red!30!}No \\    \hline
   
    \end{tabular}
    \label{tab:previousdesign}
	\vspace{-0.5cm}
\end{table}

\section{Conclusion and Future Work} \label{conclusion}
We presented a novel tip mount for transporting sensors and tools with soft growing robots that overcomes some limitations of previous tip mount designs and is able for the first time to exert significant pulling force on the environment while retracting, as well as to retrieve and deliver objects. We also analyzed the four previous soft growing robot tip mount designs in comparison with our current design. 

Future work on our current tip mount will include the development of a more robust control scheme that adds encoders to the retraction device, as well as the development of a wire management scheme for the retraction device wires. We will also improve the tip mount design limitations, including reducing weight, reducing sliding along the environment, and allowing body shrinking through apertures. 

With the successful development of methods to apply force to the environment, we are encouraged to pursue the development of soft growing robots as true manipulators that are able to move payloads with precision through a large workspace. This requires development of stronger and higher-curvature actuators for soft growing robots, as well as the ability to control and sense stiffness, shape, and force application.

\bibliographystyle{IEEEtran}
\bibliography{references}
\end{document}